\documentclass{article}

\usepackage{booktabs} 

\usepackage{microtype}
\usepackage{graphicx}
\usepackage{subcaption}
\usepackage{booktabs} 

\usepackage{graphicx}
\usepackage{amsmath}
\usepackage{amssymb}
\usepackage{bbm}
\usepackage{algorithm}
\usepackage{algorithmic}
\usepackage{color}
\usepackage{wasysym}
\usepackage{commath}
\usepackage{mathtools}
\usepackage{pdfpages}

\let\norm\undefined 
\DeclarePairedDelimiter\norm{\lVert}{\rVert}
\renewcommand\thesection{\Alph{section}}

\usepackage{hyperref}

\usepackage[accepted]{icml2019}

\icmltitlerunning{TapNet: Neural Network Augmented with Task-Adaptive Projection for Few-Shot Learning}

\begin{document}

\twocolumn[
\icmltitle{TapNet: Neural Network Augmented with Task-Adaptive Projection for Few-Shot Learning }



\icmlsetsymbol{equal}{*}

\begin{icmlauthorlist}
\icmlauthor{Sung Whan Yoon}{kaist}
\icmlauthor{Jun Seo}{kaist}
\icmlauthor{Jaekyun Moon}{kaist}

\end{icmlauthorlist}

\icmlaffiliation{kaist}{School of Electrical Engineering, Korea Advanced Institute of Science and Technology (KAIST), Daejeon, Korea}

\icmlcorrespondingauthor{Sung Whan Yoon}{shyoon8@kaist.ac.kr}

\icmlkeywords{Machine Learning, ICML}

\vskip 0.3in
]
{\large }


\printAffiliationsAndNotice{}  

\begin{abstract}
	
Handling previously unseen tasks after given only a few training examples continues to be a tough challenge in machine learning. We propose TapNets, neural networks augmented with task-adaptive projection for improved few-shot learning. Here, employing a meta-learning strategy with episode-based training, a network and a set of per-class reference vectors are learned across widely varying tasks. At the same time, for every episode, features in the embedding space are linearly projected into a new space as a form of quick task-specific conditioning. The training loss is obtained based on a distance metric between the query and the reference vectors in the projection space. Excellent generalization results in this way. When tested on the Omniglot, \textit{mini}ImageNet and \textit{tiered}ImageNet datasets, we obtain state of the art classification accuracies under various few-shot scenarios.
	
\end{abstract}

\section{Introduction}
\label{intro}

Few-shot learning promises to allow machines to carry out tasks that are previously unencountered, using only a small number of relevant examples. As such, few-shot learning finds wide applications, where labeled data are scarce or expensive, which is far more often the case than not. Unfortunately, despite immense interest and active research in recent years, few-shot learning remains an elusive challenge to machine learning community. For example, while deep networks now routinely offer near-perfect classification scores on standard image test datasets given ample training, reported results on few-shot learning still fall well below the levels that would be considered reliable in crucial real world settings. 

One popular way of developing few-shot learning strategies is to take a meta-learning perspective coupled with episodic training \cite{MN,Ravi, survey2}. Meta-learning seems to convey somewhat different meanings to different people, but none would disagree that it is about learning a general strategy to learn new tasks \cite{survey1}. Episodic training refers to a training method in which widely varying tasks (or episodes) are presented to the learning model one by one, with each episode containing only a few labeled examples. The repetitive exposure to previously unseen tasks, each time with low samples, during this initial learning or meta-training stage seems to provide a viable option for preparing the learner for quick adaptation to new data \cite{MN}. 

Among the well-known approaches in this direction are the metric-based learners like Matching Networks \cite{MN} and Prototypical Networks \cite{PN}. These methods all incorporate non-parametric, distance-based learning, where embedding space is trained to 
minimize a relevant distance metric across episodes before stabilizing to perform actual few-shot classification. 
Matching Networks train separate networks to process labeled samples and query samples, and utilize each labeled sample in the embedding space as reference points in classifying the query samples. 
Prototypical Networks employ only one embedding network with its per-class centroids used as classification references 
in the embedding space. 
Based on learning only a single feed-forward feature extractor, Prototypical Networks offer a surprisingly good ability to generalize to new tasks, as an inductive bias seems to settle in somehow via episodic training.

We are also interested in distance-based learning with no fine-tuning of parameters beyond the episodic meta-training stage. 
Relative to prior work, the unique characteristic of our method is in explicit task-dependent conditioning via linear projection of embedded features. Once the neural network outputs are projected into a new space, classification is done there based on distances from per-class reference vectors. Both the neural network and the reference vectors are learned across the sequence of episodes reflecting widely varying tasks, while the projected classification space is constructed anew specific to each episode. The projection to an alternative classification space is done via linear nulling of errors between the embedded features and the per-class references. Unlike in \cite{MN} and \cite{PN}, class-representing vectors in our scheme are not the outputs of an embedding function. Rather, the references in our method are a simple set of stand-alone vectors not directly coupled to input images, although they are updated for each episode based on distances from the embedded query images projected in the classification space.   

The combination of across-task learning of the network and per-class reference vectors with a quick task-adaptive conditioning of classification space allows excellent generalization. Extensive testing on the Omniglot, \textit{mini}ImageNet and \textit{tiered}ImageNet datasets show that the proposed network augmented with task-adaptive projection (TapNet) yields state of the art few-shot classification accuracies. 

\begin{figure*}
\centering
\includegraphics[width=0.90\textwidth]{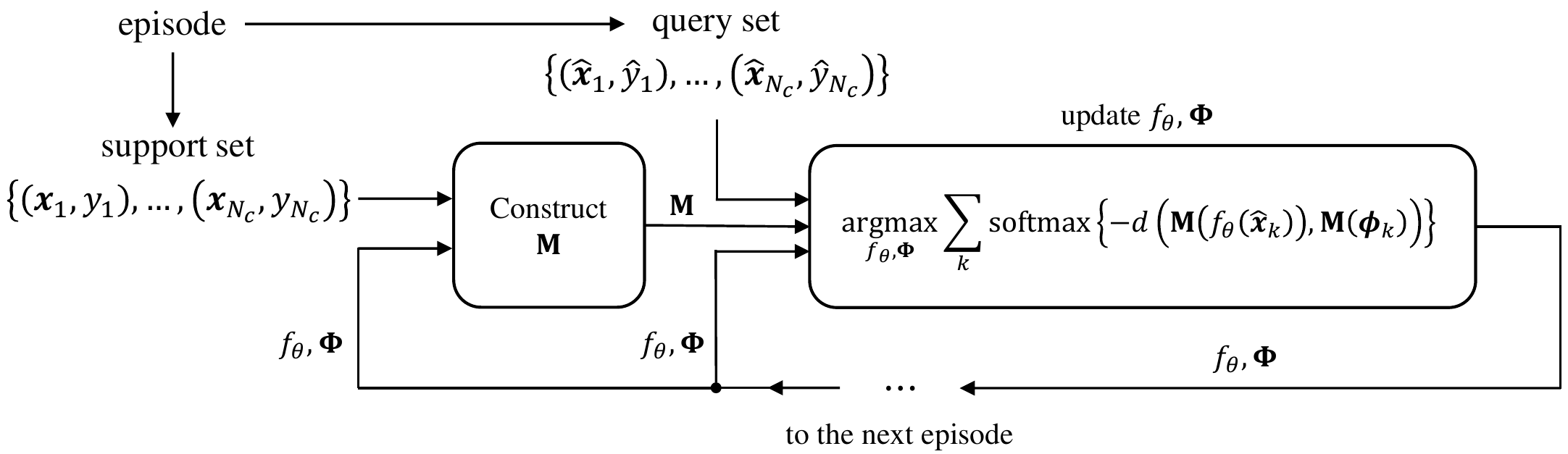}
\caption{TapNet learning process}
\label{fig:opt}
\end{figure*}

\section{Task-Adaptive Projection Network}
\subsection{Model Description}
TapNet consists of three key elements: an embedding network $f_{\theta}$, a set of per-class reference vectors $\mathbf{\Phi}$, and the task-dependent adaptive projection or mapping $\mathbf{M}$ of embedded features to a new classification space. $\mathbf{\Phi}=[\boldsymbol{\phi}_{1};\cdots;\boldsymbol{\phi}_{N_{c}}]$ is a matrix whose $k^{\text{th}}$ row is the per-class reference (row) vector $\boldsymbol{\phi}_{k}$. $\mathbf{M}$ denotes projection or mapping, but sometimes would mean the projection space itself. 
See Fig. \ref{fig:opt}, where a new episode is being presented to the model during the sequential episodic training process. An episode consists of a support set of images/labels $\{(\mathbf{x}_1, y_1),...,(\mathbf{x}_{N_c}, y_{N_c})\}$ as well as a query set $\{(\hat{\mathbf{x}}_1, \hat{y}_1),...,(\hat{\mathbf{x}}_{N_c}, \hat{y}_{N_c})\}$. For clear illustration, there is only one image/label pair for each of $N_c$ given classes, in either set here.  

Given the new support set, as well as $f_{\theta}$ and $\mathbf{\Phi}$ learned 
through the last episode stage, a projection space $\mathbf{M}$ is first constructed such that the embedded feature vectors $f_{\theta}(\mathbf{x}_k)$'s and the class reference vectors $\boldsymbol{\phi}_{k}$'s with matching labels align closely when projected into $\mathbf{M}$. The details of projection space construction will be given shortly.  

The network  $f_{\theta}$ and the reference set $\mathbf{\Phi}$ are in turn updated according to softmax based on Euclidean distance 
$\mathit{d}(\cdot,\cdot)$ between the mapped query image and reference vectors: 
\begin{align}
&\text{softmax}\Big\{-d(\mathbf{M}(f_{\theta}(\hat{\mathbf{x}}_k)),\mathbf{M}(\boldsymbol{\phi}_{k}))\Big\} \nonumber\\
&= \frac{ \exp \Big\{-\mathit{d}(\mathbf{M}(   f_{\theta}(\hat{\mathbf{x}}_k)   ),    \mathbf{M}(\boldsymbol{\phi}_{k})    ) \Big\}  }
{\sum_{l} \exp \Big\{-\mathit{d}(\mathbf{M}(f_{\theta}(\hat{\mathbf{x}}_k)),\mathbf{M}(\boldsymbol{\phi}_{l}))  \Big\}  }
\end{align}
which is averaged over all classes $k$. 
Here, $\mathbf{M(\mathbf{z})}$ denotes projection of row vector $\mathbf{z}$, and all vectors including the embedded feature vector $f_{\theta}(\mathbf{x})$ of input $\mathbf{x}$ are assumed to be row vectors, unless specified otherwise. 
The updated  $f_{\theta}$ and $\mathbf{\Phi}$ are passed to the next episode processing stage. 
The projection and parameter updates continue for each episode until all given episodes are exhausted. 

Come the few-shot test stage, again a projection space is computed to align the embedded features of the presented shots appearing at the output of the network with the references, both of which are now fixed after having learned throughout the episodic meta-training process. The query image is finally compared with the references in the projection space for final classification.

In summary, the embedder  $f_{\theta}$ and the per-class reference vectors $\mathbf{\Phi}$ are learned across varying tasks (episodes)
while the projection space  $\mathbf{M}$ is built specific to the given task, providing a quick task-dependent conditioning. This combination results in an excellent ability to generalize to new data, as extensive experimental results will verify shortly.

\subsection{Construction of Task-Adaptive Projection Space}
Finding the mapping function or projection space $\mathbf{M}$ is based on removing misalignment between the task-embedded features and the references. 
To handle general cases with multiple example images per class, let $\mathbf{c}_{k}$ be the per-class average of the embedded features for class $k$ corresponding to the images out of the support set. 

We wish to find a mapper $\mathbf{M}$ such that $\mathbf{c}_{k}$ and the matching reference vector $\boldsymbol{\phi}_k$ are highly aligned in the mapped space. At the same time, it would be beneficial to make $\mathbf{c}_{k}$ and the non-matching weights $\boldsymbol{\phi}_l$ for all $l\neq k$ well-separated in the same space. It turns out that a simple linear projection that does not require any learning offers an effective solution. The idea is to find a projection space where $\mathbf{c}_{k}$ aligns with a modified vector
\begin{align}
\tilde{\boldsymbol{\phi}}_k= \boldsymbol{\phi}_k-\frac{1}{N_c-1}\sum_{{\scriptstyle l\neq k }}\boldsymbol{\phi}_{l}
\end{align}
where the factor $1/(N_c-1)$ provides natural normalization reflecting the number of non-matching vectors. This is to say that given the error vector defined as 
\begin{equation}
\boldsymbol{\epsilon}_{k}=\frac{\tilde{\boldsymbol{\phi}}_{k}}{\norm{\tilde{\boldsymbol{\phi}}_{k}}} - \frac{\mathbf{c}_{k}}{\norm{\mathbf{c}_{k}}},
\end{equation}
$\mathbf{M}$ can be found such that the projected error vector $\boldsymbol{\epsilon}_{k}\mathbf{M}$ is zero for every $k$, where
$\mathbf{M}$ is now a matrix whose columns span the projection space. In other words, $\mathbf{M}$ is found by a linear nulling of 
errors $\boldsymbol{\epsilon}_{k}$. $\tilde{\boldsymbol{\phi}}_{k}$ and $\mathbf{c}_{k}$ are normalized to remove power imbalance between them.
Formally, we express
\begin{align}
\mathbf{M} = \text{null}_{D}\big( [ \boldsymbol{\epsilon}_{1};\cdots;\boldsymbol{\epsilon}_{N_{c}} ] \big),
\end{align}
where $D$ is the column dimension of $\mathbf{M}$.  
A well-known solution is through the singular value decomposition (SVD) of the matrix $[ \boldsymbol{\epsilon}_{1};\cdots;\boldsymbol{\epsilon}_{N_{c}} ]$, namely, by taking $D$ of the right singular vectors of the matrix, from index $N_c+1$ through $N_c+D$. With $L$ being the length of $\boldsymbol{\epsilon}_{k}$, if $L \geq N_{c}+D$, then the projection space $\mathbf{M}$ with dimension $D$ always exists.

Note that $D$ can be set less than $L-N_{c}$, indicating a possibility of significant dimension reduction. Our empirical observation suggests that a 
dimension reduction sometimes actually improves few-shot classification accuracies. Note that for SVD of an $(n\times m)$ matrix with $n\leq m$, computational complexity is $O(mn^{2})$. The required SVD computational complexity for obtaining our projection is thus $O(LN_{c}^{2})$, which is small compared to typical model complexity.

We also remark that because the solution to linear nulling as formulated above exists irrespective of particular labeling of $\boldsymbol{\phi}_{k}$'s, we do not need to relabel the reference vectors in every episode; the same label sticks to each reference vector throughout the episodic training phase.  


\subsection{Training}
As mentioned, training of the embedding network $f_{\theta}$ and the reference vectors $\mathbf{\Phi}$ is done via episodic training, following \cite{MN}.
The detailed steps of learning TapNets is provided in Algorithm \ref{alg}. 

For each training episode, $N_{c}$ classes are randomly chosen from the training set of a given dataset. 
Then, for each class, $N_{s}$ labeled samples are randomly chosen as the support set $S_{k}$, and  $N_{q}$ labeled samples are chosen as the query set $Q_{k}$, without any overlapping samples between the two sets.
With the support set $S_{k}$, the average network output vector $\mathbf{c}_{k}$ is obtained for each class (in line 5). Based on the per-class average network output vectors, error vectors are obtained for all classes (in line 6)
without any relabeling on the reference vectors.
Then the projection space $\mathbf{M}$ is computed as a null-space of the error signals, as explained in the prior subsection.
For each query input, the Euclidean distances to the reference vectors in the projection space $\mathbf{M}$ are measured, and the training loss is computed using these distances.
The average training loss is obtained over all $N_{q}$ query inputs for each of $N_{c}$ given classes (in line 11 to 14).
The learnable parameters $\theta$ of the embedding network and the references $\boldsymbol{\Phi}$ are now updated based on the average training loss (in line 16). This process gets repeated for every remaining episode with new classes of images and queries. 

Following \cite{PN}, we also use a larger number of classes than $N_c$ during episodic training, for improved performance. For example, 20-way classification is used during episodic learning or meta-training whereas 5-way classification is done in the final few-shot learning and testing. In this case, a question remains as to how 
5 reference vectors are chosen in the few-shot stage for linear projection out of 20 references that have been trained. For this, we first obtain 
the average vectors $\mathbf{c}_{k}$ for the 5 given classes and then select the 5 reference vectors closest to the $\mathbf{c}_{k}$'s and also relabel them accordingly before the linear projection is carried out. Notice that with the higher-way training employed, TapNets can easily handle cases where the number of classes for actual few-shot classification is not known in advance.


\begin{algorithm*}[h]
	\caption{Episodic learning is done by $N_{E}$ episodes. Each episode $E_{i}$ consists of $N$ (image, label) pairs. These $N$ shots are composed of $N_{s}$ support images/labels and $N_{q}$ queries from each of $N_{c}$ given classes and $N = N_c(N_s + N_q)$. $L_{\text{train}}$ is the loss for training learnable parameters. The Euclidean distance between two vectors is denoted as $d(\cdot,\cdot)$. $D$ is the dimensionality of projection space $\mathbf{M}$.}
	\label{alg}
	\textbf{Input}: Training set $E = \left\{E_1, ... ,E_{N_E} \right\}$ where $E_i$ is an episode of $N$ image/label pairs over $N_c$ classes. $E_i^{(k)}=\left\{S_k, Q_k\right\}=\left\{\{(\mathbf{x}_{k,1}, y_{k,1}),...,(\mathbf{x}_{k,N_s}, y_{k,N_s})\}, \{(\hat{\mathbf{x}}_{k,1},\hat{y}_{k,1}),...,(\hat{\mathbf{x}}_{k,N_q}, \hat{y}_{k,N_q})\}\right\}$ is a subset of $E_i$ corresponding to label $k$.
	\begin{algorithmic}[1]
		\FOR{$i$ in $ \left \{1 , ... , N_E \right \}$}
		\STATE $ \mathit{L}_{\text{train}} \leftarrow 0$
		
		\FOR{$k$ in $ \left \{1 , ... , N_c \right \}$}
		\STATE $\mathbf{c}_k \leftarrow {1 \over N_{s}} \sum_{n=1}^{N_s} {f_{\theta}(\mathbf{x}_{k,n})}$
		\STATE $\boldsymbol{\epsilon}_{k} \leftarrow \frac{\boldsymbol{\phi}_{k} - (1/(N_{c}-1))\sum_{l\neq k}{\boldsymbol{\phi}_{l}} }{\norm{\boldsymbol{\phi}_{k} - (1/(N_{c}-1))\sum_{l\neq k}{\boldsymbol{\phi}_{l}}}} - \frac{\mathbf{c}_{k}}{\norm{\mathbf{c}_{k}}}$
		
		\ENDFOR
		\STATE $ \mathbf{M} \leftarrow \text{null}_D \big( [ \boldsymbol{\epsilon}_{1};\cdots;\boldsymbol{\epsilon}_{N_{c}} ] \big) $
		\FOR{$k$ in $ \left \{1 , ... , N_c \right \}$}
		\FOR{$q$ in $ \left \{1 , ... , N_q \right \}$}
		\STATE $ \mathit{L}_{\text{train}} \leftarrow \mathit{L}_{\text{train}} + \displaystyle\frac{1}{N_{c}N_{q}}\left[d(\mathbf{M}(f_{\theta}(\hat{\mathbf{x}}_{k,q})),\mathbf{M}(\boldsymbol{\phi}_{k})) + \log \displaystyle\sum_{l}{\exp(-d(\mathbf{M}(f_{\theta}(\hat{\mathbf{x}}_{k,q})),\mathbf{M}(\boldsymbol{\phi}_{l})))}\right]$
		\ENDFOR
		\ENDFOR
		\STATE Update $ \theta,  \boldsymbol{\Phi}$ minimizing $\mathit{L}_{\text{train}}$ via optimizer
		\ENDFOR
	\end{algorithmic}
\end{algorithm*}

\section{Related Work}

\subsection{Relation to Metric-Based Meta-Learners}
Two well-known metric-based few-shot learning algorithms are Matching Networks of \cite{MN} and Prototypical Networks of \cite{PN}. Matching Networks yield decisions based on matching the output of a network driven by a query sample to the output of another network fed by labeled samples. In Matching Networks, the similarities are measured between the query output and the labeled sample outputs on two separate embeddings. The labeled sample outputs are provided as reference points for estimating the label of the query. On the other hand, Prototypical Networks are trained to minimize the distance metric between the per-class average outputs and the query output from the single embedding network.
The per-class average outputs on the embedding space work as references 
and the embedding network is trained to make a given query output stay close to the correct reference while pushing it 
away from the incorrect reference points.

Our TapNet also learns to minimize distance between the projected query and per-class references. Unlike Matching Networks and Prototypical Networks, however, there is explicit task-dependent conditioning in TapNets in the form of projection into a new classification space. Assuming one-shot support and query samples for clear illustration, the distance functions utilized by three methods are compared as:
\begin{align*}
&\text{Matching Networks:} \:\: d(f_{\theta}(\hat{\mathbf{x}}_k),g_{\phi}(\mathbf{x}_{k})) \\
&\text{Prototypical Networks:} \:\: d(f_{\theta}(\hat{\mathbf{x}}_k), f_{\theta}(\mathbf{x}_{k})) \\
&\text{TapNets:} \:\: d(\mathbf{M}(f_{\theta}(\hat{\mathbf{x}}_k)),\mathbf{M}(\boldsymbol{\phi}_{k})))
\end{align*}
where $\hat{\mathbf{x}}_k$ and $\mathbf{x}_{k}$ are the one-shot query and support samples, respectively, for class $k$. Matching Networks
use two separate embedding functions $f_{\theta}$ and $g_{\phi}$, while Prototypical Networks rely on a single embedding function. TapNets employ one embedding network but there is a learnable reference vector set $\boldsymbol{\phi}_{k}$'s as well as task-conditioning linear projection $\mathbf{M}$. Also note that the class references in Matching Networks and Prototypical Networks are embedded features themselves, but
those in TapNets are not; rather, the per-class references in TapNets are stand-alone vectors that are not directly coupled to the input images.   

Built upon a base Prototypical Network model, TADAM of \cite{TADAM} employs learned metric-scaling and additional task-dependent conditioning in the form of element-wise scaling and shifts for the feature vectors of component convolutional layers, similar to \cite{FILM}. But this conditioning requires learning of extra fully-connected networks, whereas the task-conditioned $\mathbf{M}$ in TapNets is computed directly from the embedded features of the new task and the up-to-date references. A metric-based learner utilizing a nonlinear distance metric was also suggested in \cite{Yang}, where the distance metric itself is learned together with the embedding network.  

\subsection{Relation to Memory-Augmented Neural Network}

While our TapNet is close in spirit to the metric-based learners in the form of Matching Networks and Prototypical Networks as described above, it also bears a surprisingly close connection to the memory-augmented neural network (MANN) of \cite{MANN}. MANN 
utilizes an external memory module with its contents rapidly adapting to new samples while the read and write weights learned across episodes. 
The explicit form of the memory module is an external matrix array $\mathbf{M}_e$ designed to store information extracted by the controller network for a given episode. The memory read output vector $\mathbf{r}$ is seen as a weighted linear combination of the columns of $\mathbf{M}_e$: 
$\mathbf{r} \leftarrow \sum_i \mathit{w}_i^r \mathbf{m}_i$. The read weight is proportional to the cosine similarity of the column $i$ of the memory with the controller network output or key vector $\mathbf{k}$:  
\begin{align*}
\mathit{w}_i^r \leftarrow \frac{ \exp\left(\mathit{K}(\mathbf{k},\mathbf{m}_i)\right)}{\sum_{j \neq i} \exp\left(\mathit{K}(\mathbf{k},\mathbf{m}(j))\right)} 
\end{align*}
where $\mathit{K}(\cdot,\cdot)$ is the cosine correlation of two vectors. Approximating the exponentiation by a linear function, i.e., $\exp(\mathit{K}(\mathbf{k},\mathbf{m}_i)) \simeq 1+\mathit{K}(\mathbf{k},\mathbf{m}_i)$, and further dropping the class-independent constant,  
we can write 
\begin{gather*}
\mathbf{w}^r \sim  \mathbf{M}_e^{T} \mathbf{k}^{T}
\end{gather*}
where $\mathbf{w}^r$ is a column vector of the read weights. 
Eventually, we can approximate the read output $\mathbf{r}$ as
\begin{equation}
\mathbf{r}  \sim \mathbf{M}_e\mathbf{M}_e^{T} \mathbf{k}^{T}.\notag
\end{equation}
In arriving at the final decision, this memory read output is multiplied by the matrix $\mathbf{W}=[\mathbf{w}_1;\cdots;\mathbf{w}_{N_{c}}]$ whose row vectors are learnable per-class weights:
\begin{equation}\label{eq:mann}
\mathbf{W}\mathbf{M}_e\mathbf{M}_e^{T} \mathbf{k}^{T}
\end{equation}
While the inference from the direct branch of the hidden state of the long short-term memory (LSTM) is also used in the MANN, we consider only the inference from the read vector $\mathbf{r}$ which utilizes information from the external memory.
The resulting vector of (\ref{eq:mann}) is the inner-product similarities between the read output $\mathbf{r}$ and weights $\mathbf{W}$. 

Going back to our TapNet, if we were to use the cosine similarity instead of Euclidean distance, the distance profile between the query and the per-class references would be  
 $\mathbf{\Phi}\mathbf{M}\mathbf{M}^{T}(f_\theta(\mathbf{x}))^{T}$, which is essentially the same as (\ref{eq:mann}) of MANN, with 
  $\mathbf{\Phi}$ playing the same role as $\mathbf{W}$, given that the key vector $\mathbf{k}$ of MANN is the same as the embedded feature of the input image for TapNets. Very interestingly, an important implication here is that the external memory array $\mathbf{M}_{e}$ of MANN can be interpreted as a kind of task-adaptive projection space where the similarities between the query key $\mathbf{k}$ and the weights $\mathbf{W}$ are measured. 
  
  \subsection{Other Types of Optimizers}
  
  There are other types of approaches often referred to as the optimization-based meta-learners. 
  They aim to optimize the embedding network quickly so that the fine-tuned network successfully adapts to the given task. 
  The meta-learner LSTM of \cite{Ravi} is one such approach, where an LSTM  \cite{LSTM} is trained to optimize another learner, which performs actual classification. There, the parameters of the few-shot learner are first set to the memory state of the LSTM, and then quickly learned based on the memory update rule of the LSTM, effectively capturing the knowledge from small samples. 
  
  The model-agnostic meta-learner (MAML) of \cite{MAML} sets up the model for easy fine-tuning so that a small number of gradient-based updates allows the model to yield a good generalization to new tasks. The fine-tuning method of MAML has been incorporated into many other schemes, such as Reptile of \cite{REPTILE} and Platipus of \cite{PMAML}. 
  Very recently, meta-learners with latent embedding optimization (LEO) of \cite{LEO} have been introduced that attempt at task-dependent 
  initialization of the model parameters with additional fine-tuning of the parameters in a low-dimensional latent space. Also, separate pre-training is required over the same training dataset in order for LEO to work properly.
  There also exist other meta-learners employing different forms of task-conditioning \cite{adares}. 
  A method dubbed the simple neural attentive meta-learner (SNAIL) combines an embedding network with temporal convolution and soft attention to draw from past experience while attempting precision access at the same time 
  \cite{SNAIL}.  
  
\section{Experiment Results}

\begin{table*}[h]
  \caption{Few-shot classification accuracies for 20-way Omniglot and 5-way \textit{mini}ImageNet}
  \label{acc_table1}
  \centering
  \begin{tabular}{l||cc||cc}
    \toprule  
    & \multicolumn{2}{c}{\textbf{20-way Omniglot}} & \multicolumn{2}{c}{\textbf{5-way \textit{mini}ImageNet}} \\
    \cmidrule{2-5}
    \textbf{Methods}    & 1-shot & 5-shot	& 1-shot & 5-shot  \\
    \midrule
    \textbf{Matching Nets} \cite{MN}  & 88.2\%	& 97.0\% & 43.56 $\pm$ 0.84\%  & 55.31 $\pm$ 0.73\%   \\
    \textbf{MAML} \cite{MAML} & 95.8\%	& 98.9\% & 48.70 $\pm$ 1.84\%  & 63.15 $\pm$ 0.91\%	 \\
    \textbf{Prototypical Nets} \cite{PN} & 96.0\%	& 98.9\% & 49.42 $\pm$ 0.78\%  & 68.20 $\pm$ 0.66\%	  \\
    \textbf{SNAIL} \cite{SNAIL}  & 97.64\% & 99.36\% & 55.71 $\pm$ 0.99\%  & 68.88 $\pm$ 0.92\%  \\
    \textbf{adaResNet} \cite{adares} & 96.12\% & 98.48\% & 56.88 $\pm$ 0.62\% & 71.94 $\pm$ 0.57\% \\
    \textbf{Transductive Propagation Nets} \cite{Liu} & - & - & 55.51 $\pm$ 0.86\% & 69.86 $\pm$ 0.65\% \\
    \textbf{TADAM-}$\boldsymbol{\alpha}$ \cite{TADAM} & - & - &56.8 $\pm$ 0.3\% & 75.7 $\pm$ 0.2\% \\ 
    \textbf{TADAM-TC} \cite{TADAM}  & - & - &58.5 $\pm$ 0.3\% & \textbf{76.7} $\pm$ \textbf{0.3}\% \\
    \textbf{TapNet} (Ours)  & \textbf{98.07}\% & \textbf{99.49}\%  &\textbf{61.65} $\pm$ \textbf{0.15}\%  & \textbf{76.36} $\pm$ \textbf{0.10}\%\\
    \bottomrule
  \end{tabular}
\end{table*}

\begin{table*}[h]
  \caption{Few-shot classification accuracies for 5-way \textit{tiered}ImageNet}
  \label{acc_table2}
  \centering
  \begin{tabular}{l||cc}
    \toprule  
    &  \multicolumn{2}{c}{\textbf{5-way \textit{tiered}ImageNet}} \\
    \cmidrule{2-3}
    \textbf{Methods} 	& 1-shot & 5-shot  \\
    \midrule
    \textbf{MAML} (as evaluated in \cite{Liu})  & 51.67 $\pm$ 1.81\%  & 70.30 $\pm$ 1.75\%   \\
    \textbf{Prototypical Nets} (as evaluated in \cite{Liu}) & 53.31 $\pm$ 0.89\%  & 72.69 $\pm$ 0.74\%	 \\
    \textbf{Relation Nets} (as evaluated in \cite{Liu}) & 54.48 $\pm$ 0.93\%  & 71.31 $\pm$ 0.78\%	  \\
    \textbf{Transductive Propagation Nets} \cite{Liu}  & 59.91 $\pm$ 0.94\%  & 73.30 $\pm$ 0.75\%  \\
    \textbf{TapNet} (Ours)   &\textbf{63.08} $\pm$ \textbf{0.15}\%  & \textbf{80.26} $\pm$ \textbf{0.12}\%\\
    \bottomrule
  \end{tabular}
\end{table*}

\subsection{Datasets}
\textbf{Omniglot} \cite{Omniglot} is a set of images of 1623 handwritten characters from 50 alphabets with 20 examples for each class. We have used 28$\times$28 downsized grayscale images and introduced class-level data augmentation by random angle rotation of images in multiples of 90$^\circ$ degrees, as done in prior works \cite{MANN,PN,MN}. 1200 characters are used for training and test is done with the remaining characters.

\textbf{\textit{mini}ImageNet} \cite{MN} is a dataset suggested by Vinyals et al. for few-shot classification of colored images. It is a subset of the ILSVRC-12 ImageNet dataset \cite{Imagenet} with 100 classes and 600 images per class. We have used the splits introduced by Ravi and Larochelle \cite{Ravi}. For our experiment, we have used 84$\times$84 downsized color images with a split of 64 training classes, 16 validation classes and 20 test classes.

\textbf{\textit{tiered}ImageNet} \cite{MRen} is a dataset suggested by Ren et al. It is a larger subset of the ILSVRC-12 ImageNet dataset with 608 classes and 779,165 images in total. The classes in \textit{tiered}ImageNet are grouped into 34 categories corresponding to higher-level nodes in the ImageNet hierarchy curated by human \cite{Deng}. These categories are split into 20 training, 6 validation and 8 test categories, and the training, validation and test sets contain 351, 97 and 160 classes, respectively. The split of \textit{tiered}ImageNet ensures that the classes in the training set are distinct from those in the test set, possibly resulting in more realistic classification scenarios.

\begin{figure*}[!htbp]
	\centering
	\includegraphics[width=120mm]{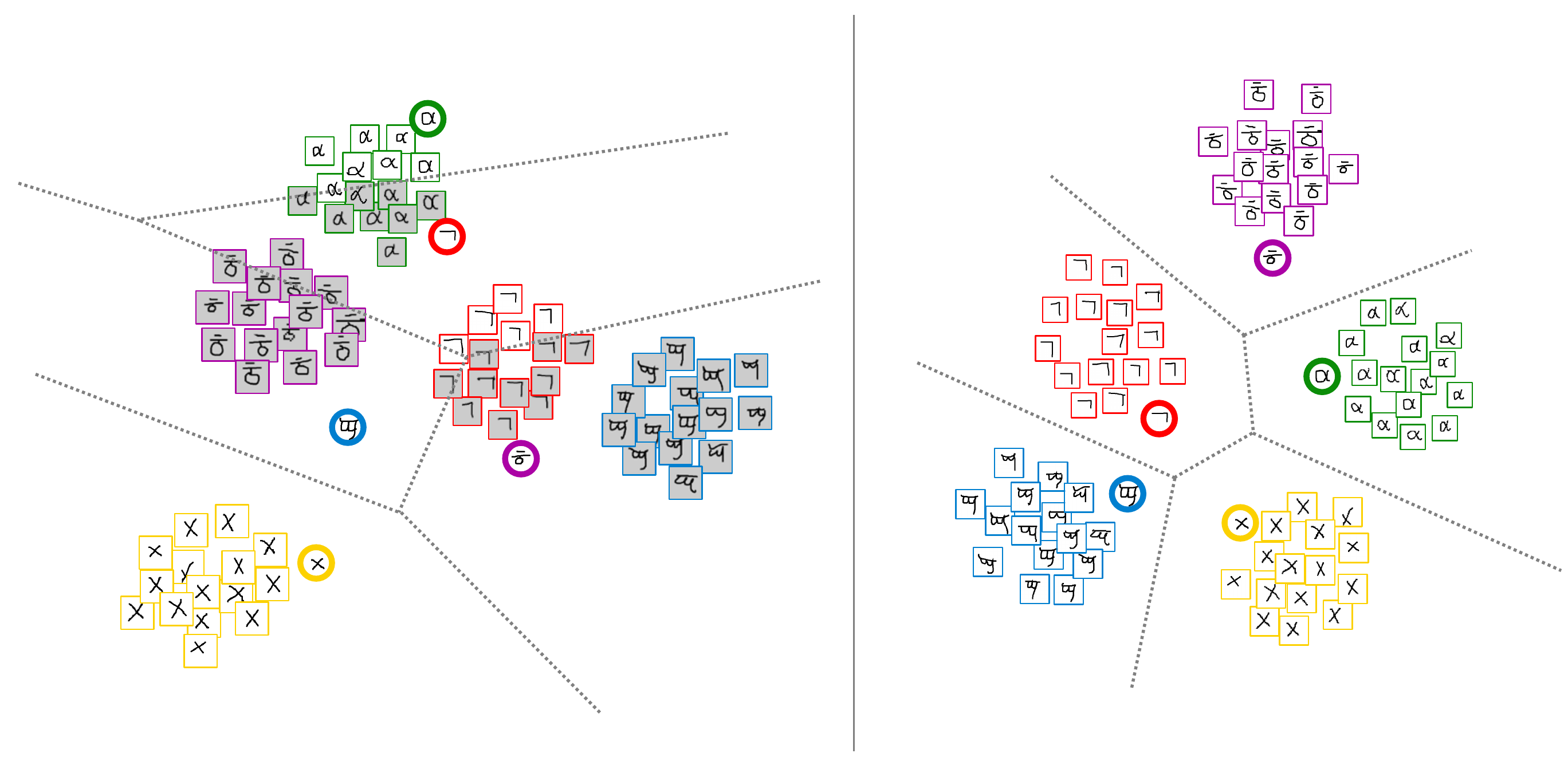}
	\caption{t-SNE visualization of the network embedding space (left) and projection space (right) }
	\label{fig:tSNE_diagram}
\end{figure*}

\subsection{Experimental Settings}


For all our experiments here, we employ ResNet-12 \cite{Resnet} as the embedding network (results with smaller networks are also available as discussed in Supplementary Material).
 ResNet-12 has four residual blocks, each of which contains three 3$\times$3 convolution layers and convolutional shortcut connection. Each convolution layer is followed by a batch normalization layer and ReLU activation. A 2$\times$2 max-pooling is applied at the end of each residual block. At the top of the stack of residual blocks, we also apply a global average-pooling to reduce feature dimensionality. We use different numbers of channels for the three datasets. For 5-way \textit{mini}ImageNet and 5-way \textit{tiered}ImageNet, the number of channels starts with 64, and doubles after the max-pooling is applied. For 20-way Omniglot classification, the number of channels starts with 64 and then increases for the subsequent blocks, although less channels are employed at later blocks than the \textit{mini}ImageNet and \textit{tiered}ImageNet cases.

The Adam optimizer \cite{Adam} with an optimized learning-rate decay is employed. 
For all experiments, the initial learning rate is $10^{-3}$. In the 20-way Omniglot experiment, the learning rate is reduced by half at every $4.0\times 10^{4}$ episodes, but for 5-way \textit{mini}ImageNet and 5-way \textit{tiered}ImageNet classification, we cut the learning rate by a factor of 10 at every $2.0\times 10^{4}$ and $4.0\times 10^{4}$ episodes, respectively,  for 1-shot experiments and every $4.0\times 10^{4}$ and $3.0\times 10^{4}$ episodes, respectively, for 5-shot experiments.

For meta-training of the network, we adopt the higher-way training of Prototypical Networks of \cite{PN}. We used 60-way episodes for 20-way Omniglot classification, and 20-way episodes for 5-way \textit{mini}ImageNet and \textit{tiered}ImageNet classification for training. In the test phase, we have to choose 20 and 5 references among 60 and 20 vectors, respectively. In selecting only a subset of reference vectors for testing purposes, relabeling is done. For each average network output chosen in arbitrary order, the closest vector among the remaining ones in $\mathbf{\Phi}$ is tagged with the matching label.
The closeness measure is the Euclidean distance in our experiments. 
After choosing the closest reference vectors, the projection space $\mathbf{M}$ is obtained for few-shot classification. The experimental results of our meta-learner in Tables \ref{acc_table1} and \ref{acc_table2} are based on 60-way initial learning for 20-way Omniglot and 20-way initial learning for 5-way \textit{mini}ImageNet and 5-way \textit{tiered}ImageNet classification.

For 20-way Omniglot classification, we used $1.0\times 10^{5}$ training episodes with 15 query samples per class. For both 5-way \textit{mini}ImageNet and 5-way \textit{tiered}ImageNet settings, $5.0\times 10^{4}$ training episodes with 8 query samples per class are used. For all experiments, we pick the best model with the highest validation accuracy during meta-training. Additional parameter settings are considered in Supplementary Material.
\subsection{Results}

In Tables \ref{acc_table1} and \ref{acc_table2}, few-shot classification accuracies on the Omniglot, \textit{mini}ImageNet and \textit{tiered}ImageNet datasets are compared\footnote{Codes are available on https://github.com/istarjun/TapNet}. The performance in the 20-way Omniglot experiment is evaluated by the average accuracy over randomly chosen $1.0\times10^{4}$ test episodes with 5 query images for each class. On the other hand, the performance in 5-way \textit{mini}ImageNet and \textit{tiered}ImageNet is evaluated by the average accuracy and a 95\% confidence interval over randomly chosen $3.0\times10^{4}$ test episodes with 15 query images for each class. 

For the 20-way Omniglot results in Table \ref{acc_table1}, TapNet shows the best performance for both 1-shot and 5-shot cases. For the 5-way \textit{mini}ImageNet results, our TapNet shows the best 1-shot accuracy and a 5-shot accuracy comparable to the best,
that of TADAM-TC, in the sense that the confidence intervals overlap. 
TADAM-$\alpha$ is a model with metric scaling but without task-conditioning, built upon Prototypical Networks with the same ResNet-12 base architecture we use. TapNet achieves a higher accuracy than TADAM-$\alpha$. TADAM-TC requires additional fully-connected layers for task-conditioning.

In Table \ref{acc_table2}, the results for \textit{tiered}ImageNet experiments are presented. We compared our method with prior work as evaluated in \cite{Liu}. Our TapNet also achieves the best performance for both 1-shot and 5-shot classification tasks with considerable margins.

We remark that although better results have been reported recently in \cite{LEO}, the base feature extractor used there is Wide ResNet-28-10, which is considerably larger - more than twice as deep and also much wider - than our base model, ResNet-12. In addition, for the method of \cite{LEO}, a separate round of pre-training is necessary over the same training set. At this point, we do not make direct performance comparison with \cite{LEO}.
Additional experimental results with varying network sizes are presented in Supplementary Material.

\begin{figure}[!h]
	\centering
	\includegraphics[width=0.40\textwidth]{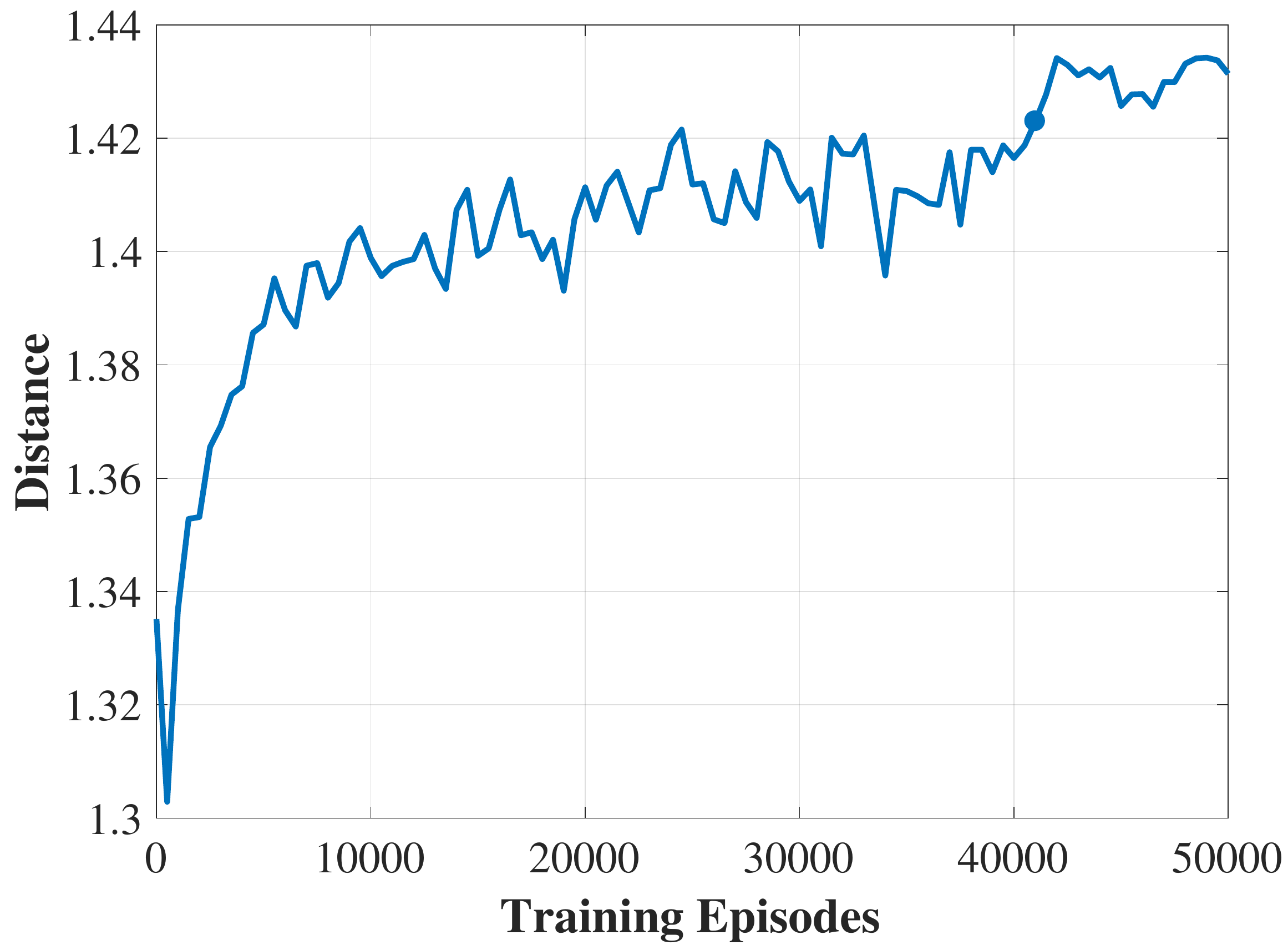}
	\caption{Minimum distance between the normalized references}
	\label{fig:dist_phi}
\end{figure}


\begin{figure*}[h]
	\centering
	\begin{subfigure}[b]{0.38\textwidth}
		\includegraphics[width=\textwidth]{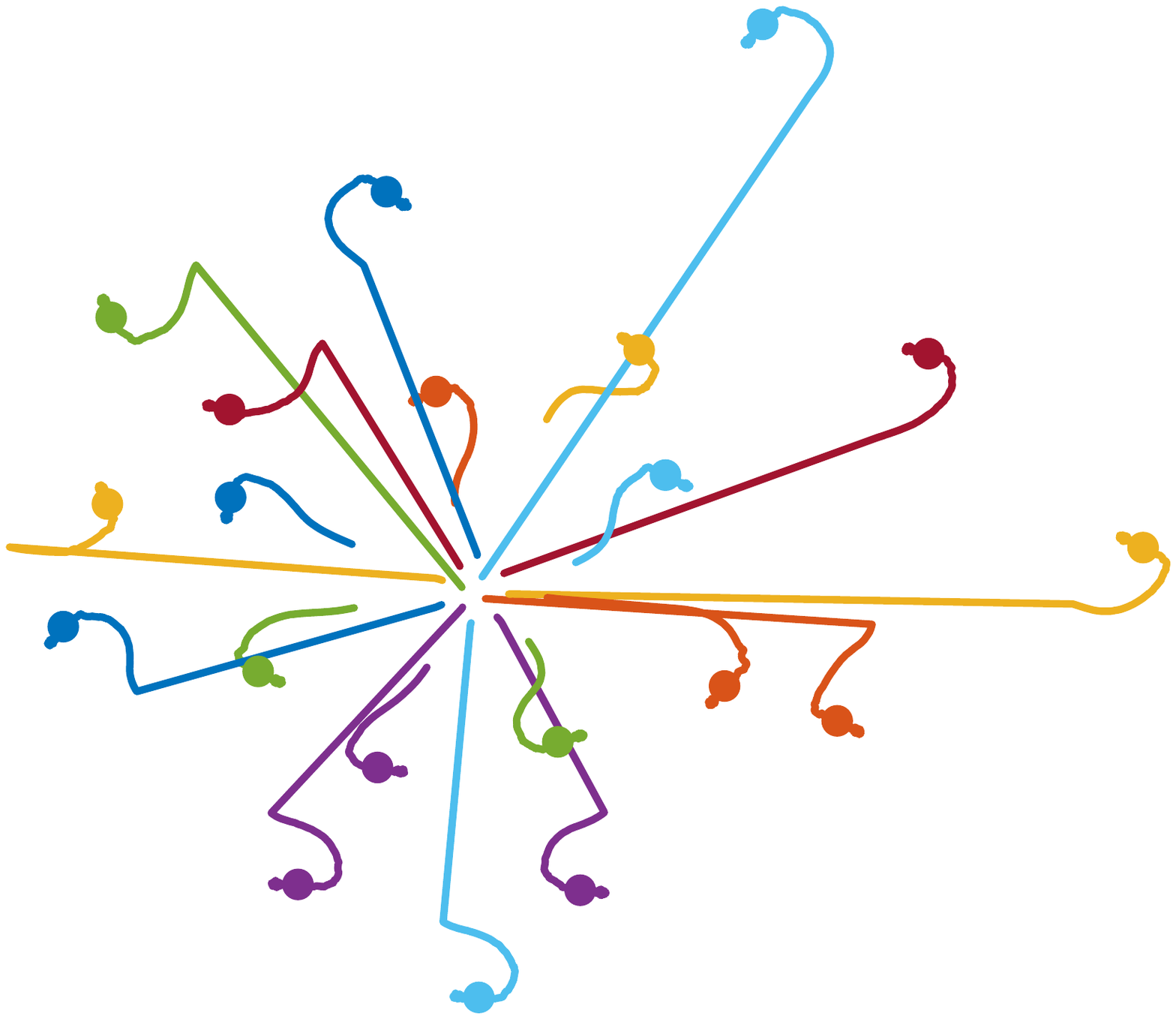}
		\subcaption{t-SNE visualization for the trajectories of $\boldsymbol{\phi}_{k}$'s}
		\label{fig:tsne_phi}
	\end{subfigure}	
	\label{fig:PhiTrain}
	\begin{subfigure}[b]{0.38\textwidth}
		\includegraphics[width=\textwidth]{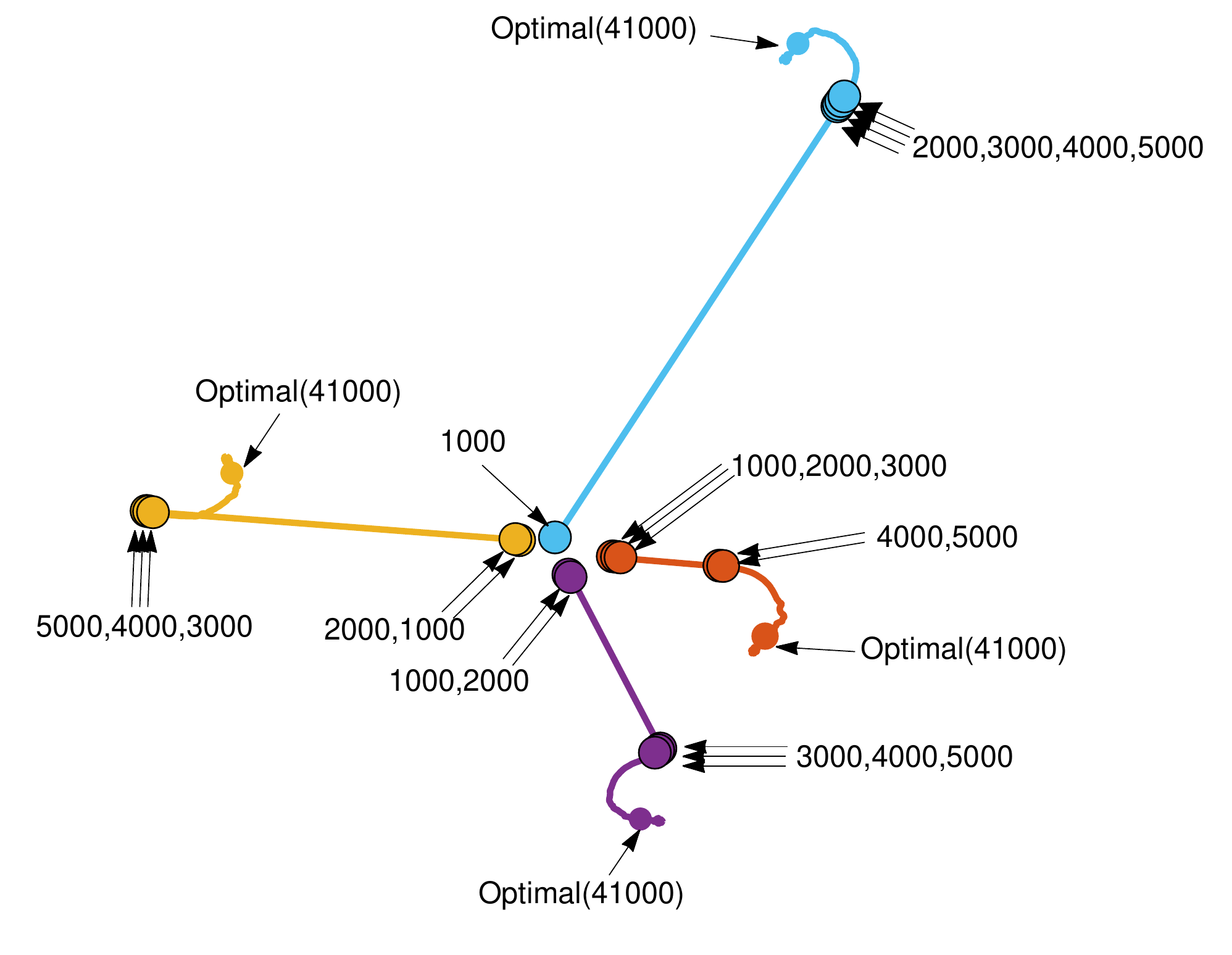}
		\subcaption{t-SNE visualization of four selected references}
		\label{fig:tsne_phi3}
	\end{subfigure}	
	\caption{Trajectories of $\boldsymbol{\Phi}$ in episodic learning}
	\label{fig:tsne_phi_full}
\end{figure*} 

\textbf{t-SNE plot.}
Fig. \ref{fig:tSNE_diagram} is a t-SNE visualization of network embedding space and projection space for TapNets trained with the Omniglot dataset. The reference vector $\boldsymbol{\phi}_{k}$'s are marked as the alphabet images in the circles. We can observe that the extracted images and the corresponding reference vectors are not located closely in the embedding space, while they lie close in the projection space. Moreover, note that the projected images are not only closely located to the matching references, but also tend to be away from non-matching references. This is due to the inclusion of the non-matching references in the modified reference vector
$\tilde{\boldsymbol{\phi}}_{k}=\boldsymbol{\phi}_{k} - (1/(N_{c}-1))\sum_{l\neq k}{\boldsymbol{\phi}_{l}}$,
utilized in defining the error vector to null in the linear projection process.

\textbf{Learning trend of references $\boldsymbol{\Phi}$.} In Figs. \ref{fig:dist_phi} and \ref{fig:tsne_phi_full}, we visualize how the labeled references $\mathbf{\Phi}$ are being optimized during the meta-training phase. 
First, Fig. \ref{fig:dist_phi} shows a plot of the Euclidean minimum distance between the normalized $\boldsymbol{\phi}_{k}$'s.
The minimum distance escalates quickly during the first $1.0\times 10^{4}$ episodes, and then increases slowly until the learning rate decay after $4.0\times 10^{4}$ episodes. This implies that the references tend to grow apart, showing improving separation over time. The solid dot indicates the moment where the model yields the highest validation accuracy.
We can develop further insights by exploring the trajectories of the vector tips of $\mathbf{\phi}_{k}$'s. Fig. \ref{fig:tsne_phi} represents a t-SNE visualization of the trajectories of 20 references.
From the random initial points which are not well-separated, the references spread out for better separation, consistent with the observation from the minimum distance plot.
In Fig. \ref{fig:tsne_phi3}, only 4 references are selected for a clearer illustration. For a given reference vector, the numbers shown by the pointing arrows represent time stamps as indicated by the number of episodes processed. As seen, there appears to be a sudden jump at some point as the vector tip grows radially, and as it reaches around the optimal point it no longer seems to move outward, tending to settle into a place.

\begin{figure}[!h]
	\centering
	\includegraphics[width=0.39\textwidth]{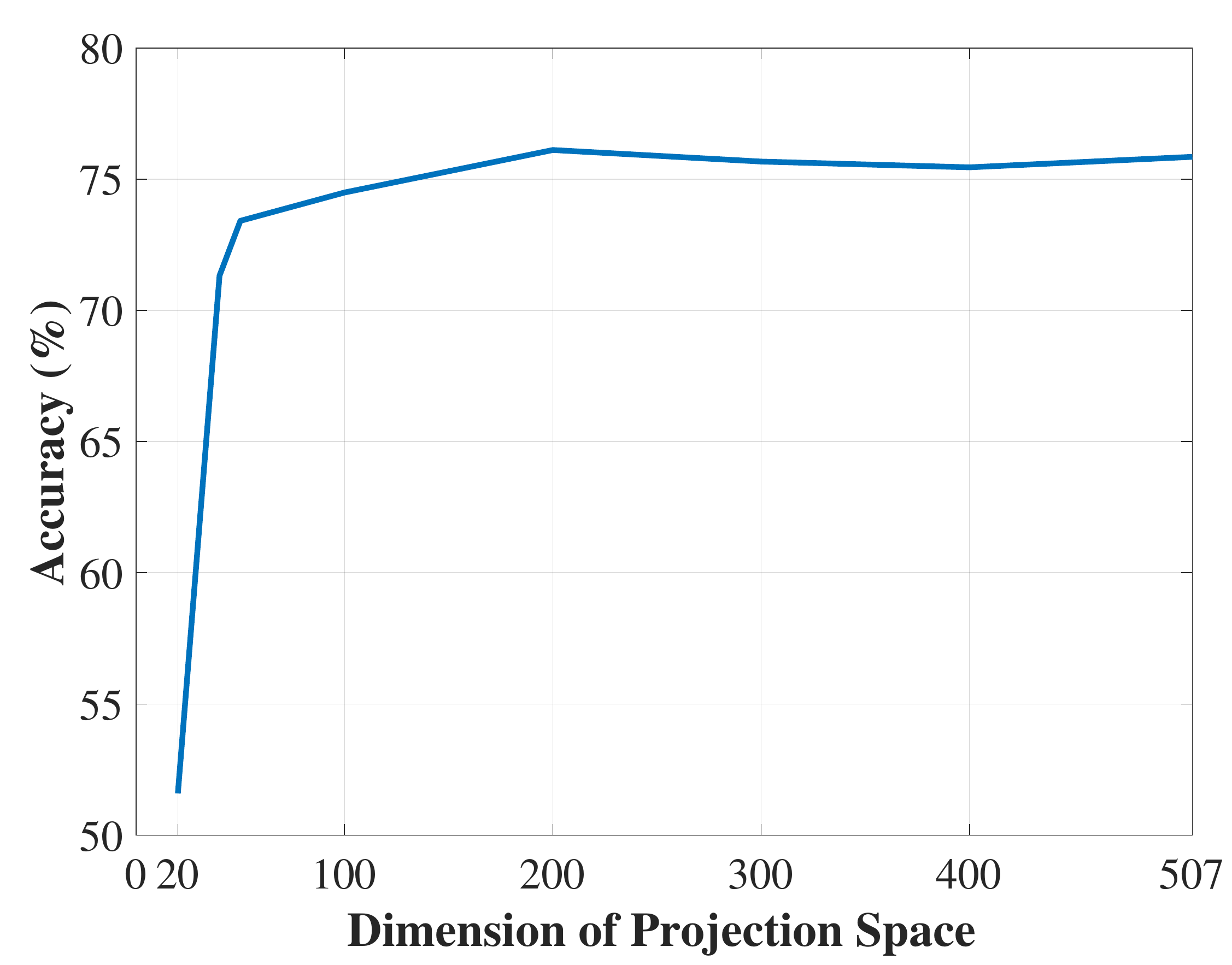}
	\caption{Test accuracies for 5-way 5-shot \textit{mini}ImageNet classification with different projection space dimensionalities}
	\label{fig:dimension}
\end{figure}

\textbf{Dimensionality of projection space.}
We study the effects of the dimensionality of projection space $\mathbf{M}$ on generalization performance. 
For the 5-way \textit{mini}ImageNet experiments, the full dimensionality $D$ of $\mathbf{M}$ is $(512-N_{c})$. This means that  $D$ is 492 for training and 507 for testing.
We carry out the 5-way 5-shot \textit{mini}ImageNet classification test with projection using various values of $D$.  
In Fig. \ref{fig:dimension}, we  observe that the test accuracy improves rapidly as dimensionality reaches around 50 and then settles down with a slight peak around 200. In our experiments we typically use full dimension with an exception of 
5-way, 5-shot \textit{mini}ImageNet classification, where $D=200$ is used, as taught by Fig. \ref{fig:dimension}. 

\section{Conclusions}
In this work, we proposed a few-shot learning algorithm aided by a linear transformer that 
performed task-specific null-space projection of the network output.
The original feature space is linearly
projected into another space where actual classification takes place.
Both the embedding network and the per-class references are learned over the entire episodes while 
projection is specific to the given episode. The resulting combination shows an excellent generalization capability to new tasks. 
State of the art few-shot classification accuracies have been observed on standard image datasets. Relationships to other metric-based meta-learners 
as well as memory-augmented networks have been explored. 


\nocite{langley00}

\section*{Acknowledgements}
This work is supported in part by the ICT R\&D program of Institute for Information \& Communications Technology Promotion (IITP) grant funded by the Korea government (MSIP) [2016-0-005630031001, Research on Adaptive Machine Learning Technology Development for Intelligent Autonomous Digital Companion].

\bibliography{ICML2019}
\bibliographystyle{icml2019}


\section*{Supplementary material}

# TapNet: Neural Network Augmented with Task-Adaptive Projection for Few-Shot Learning: Supplementary Material

**Sung Whan Yoon**[1]  **Jun Seo**[1]  **Jaekyun Moon**[1]

## 1. Architecture Details

To obtain the results presented in the main paper, we employed ResNet-12, a residual network with 12 convolutional layers, as the feature extractor of TapNets. ResNet-12 is composed of four residual blocks and four max-pooling layers. Each residual block is constructed with three layers of $3 \times 3$ convolutions, followed by a batch normalization layer and an ReLU activation function. Each residual connection consists of a $3 \times 3$ convolutional layer and a batch normalization layer, and links the input to the last activation function. At the top of the residual block stack, global average pooling is also applied to reduce dimension. The four residual blocks have 64, 128, 256 and 512 respective channels. As mentioned in the main paper, this ResNet-12 is the same embedding network used in the task-dependent adaptive metric (TADAM) scheme.

## 2. Ablation Study

In this section, we show the results of ablation studies involving the following issues: composition of training episodes, learning rate optimization, regularization hyperparameters and reference vector settings. We used the Adam optimizer for all experiments. Also, $l2$ regularization and dropout are employed.

### 2.1. Episode Composition

Fig. 1 shows test accuracies for 5-way, 5-shot *mini*ImageNet with different numbers of training classes and query samples per class. We used the Adam optimizer with an initial learning rate of $10^{-3}$ and cut the learning rate by a factor of 10 every $4.0 \times 10^4$ episodes. $l2$ regularization with a weight decay rate of $5.0 \times 10^{-4}$ is used and dropout with ratio 0.2 is applied to every output of the max-pooling layers. As summarized in Table 1, the best accuracy is obtained using

either 20 training classes with 8 query samples per class or 25 training classes with 6 query samples per class. We simply chose the 20/8 combination for TapNet experiments.

*Table 1.* Best $N_q$ for different numbers of training classes

| Number of training classes | Best $N_q$ | Accuracy |
|:---:|:---:|:---:|
| 5 | 40 | 74.19% |
| 10 | 14 | 75.73% |
| 15 | 10 | 75.40% |
| **20** | **8** | **75.86%** |
| 25 | 6 | 75.86% |
| 30 | 4 | 75.09% |

### 2.2. Hyperparameters in Experiment

Table 2 shows the hyperparameter values used for TapNet experiments. For 20-way Omniglot and 5-way *tiered*ImageNet experiments, we used step decays for the learning rate and dropout. In 5-way *mini*ImageNet experiments, $l2$ regularization is also applied in addition to dropout and learning rate decays. The dropout ratio bracket indicates the dropout ratio applied to each of the four pooling layer outputs.

### 2.3. Ablation Study for Reference Vectors

We test the 5-way, 5-shot *mini*ImageNet classification accuracy while varying the settings on the reference vectors $\Phi$. In the main paper, we use the modified reference vectors $\tilde{\phi}_k = \phi_k - \frac{1}{N_c - 1} \sum_{l \neq k} \phi_l$ when constructing the projection space, but use the original reference vectors $\phi_k$ when classifying the query samples. Using the modified references in constructing projection space gives better separation among the classes in the projection space. By constructing projection space with the original reference vectors, TapNet achieves 75.53% accuracy, a degradation from 76.36%. Also, one can think of using the modified reference vectors in place of the original reference vectors during classification of the query samples. In this case, the classification accuracy also degrades to 74.61%.

For meta-training of TapNets, we adopted the higher way training. As a result, the number of prepared reference vectors is larger than the actual number of distinct classes

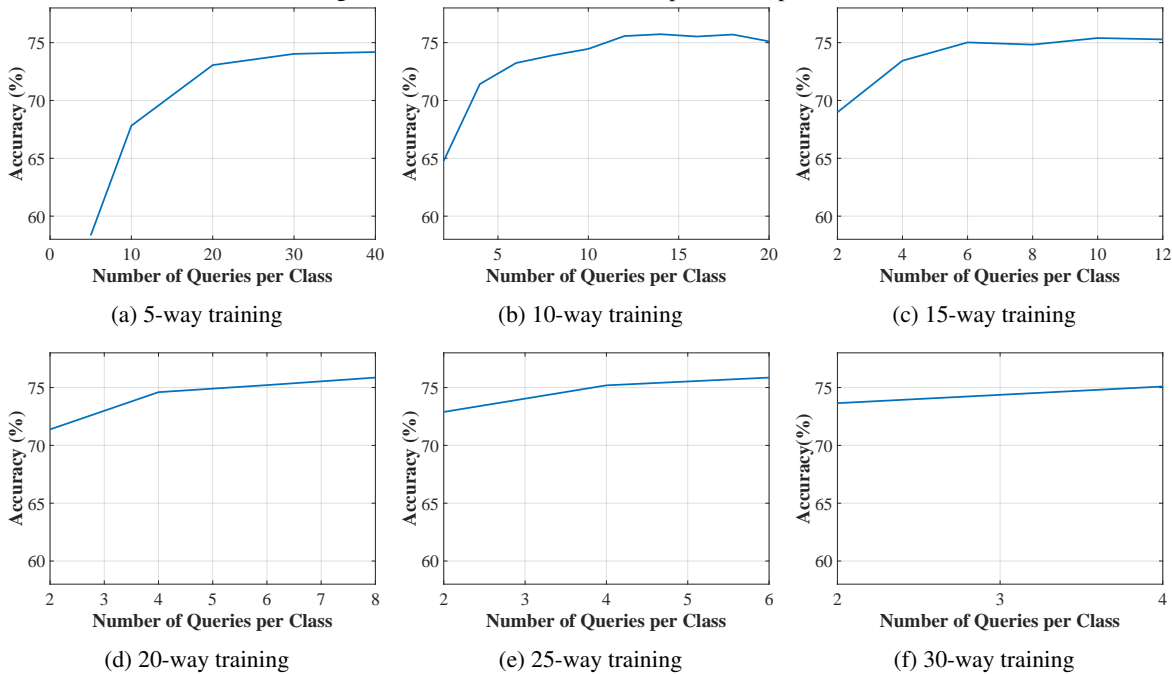

*Figure 1.* Accuracies with different episode compositions

*Table 2.* Hyperparameter settings for meta-training

| **Model** | $N_c$ | $N_q$ | learning rate | lr decay step | lr decay ratio | $l2$ decay rate | dropout ratio |
|---|---|---|---|---|---|---|---|
| **Omniglot** 1-shot case | 60 | 15 | 1e-3 | 40000 | 0.5 | - | [0.2,0.2,0.2,0.2] |
| **Omniglot** 5-shot case | 60 | 15 | 1e-3 | 40000 | 0.5 | - | [0.2,0.2,0.2,0.2] |
| ***mini*ImageNet** 1-shot case | 20 | 12 | 1e-3 | 20000 | 0.1 | 5e-4 | [0.2,0.2,0.2,0.2] |
| ***mini*ImageNet** 5-shot case | 20 | 8 | 1e-3 | 40000 | 0.1 | 5e-4 | [0.3,0.2,0.2,0.2] |
| ***tiered*ImageNet** 1-shot case | 30 | 8 | 1e-3 | 40000 | 0.1 | - | [0.2,0.2,0.2,0.2] |
| ***tiered*ImageNet** 5-shot case | 20 | 8 | 1e-3 | 30000 | 0.1 | - | [0.2,0.2,0.2,0.2] |

during evaluation. This leads to a study of the effect of particular reference vector selection. In the main paper, we selected those reference vectors closest to the class average vectors. We test two other possible reference selections here: one is simple random selection and the other is to select the farthest references to the class average. On the 5-way, 5-shot *mini*ImageNet classification task, random selection shows a slightly degraded accuracy of 75.78% while the farthest reference selection yields a bit more drop to 75.11%. In summary, although choosing the reference vectors closest to the class averages gives the best performance, any particular selection seems to make only a small difference.

In Section 4.3 of the main paper, we studied the learning trend of the reference $\Phi$ based on minimum distance growth as well as visualization of the vector trajectories. As easily seen from the figures, the norm of each reference vector increased as training progressed. This naturally raises a question: would initializing the references with a larger norm be beneficial? The answer, however, turned out to be no; when the reference vectors were initialized to have a norm as large as the fully meta-trained reference vectors, the performance of TapNet actually dropped slightly to 75.78%.

## 3. Experimental Results for Varying Network Sizes

We now evaluate TapNets with varying embedding network sizes. The first option is the convolutional neural network (CNN) widely used in prior works such as Matching Networks or Prototypical Networks. It is based on four convolutional blocks, each of which consists of a 3×3 convolutional layer with 64 filters, stride 1 and padding along with a batch normalization layer, a ReLU activation and a 2×2 max-pooling. This CNN is denoted as "Conv4" in Table 3. The

*Table 3.* Few-shot classification accuracies for 5-way *mini*ImageNet

| Methods | 5-way *mini*ImageNet | |
|---|---|---|
| | 1-shot | 5-shot |
| **Matching Networks** | $43.56 \pm 0.84\%$ | $55.31 \pm 0.73\%$ |
| **MAML** | $48.70 \pm 1.84\%$ | $63.15 \pm 0.91\%$ |
| **Prototypical Networks** | $49.42 \pm 0.78\%$ | $68.20 \pm 0.66\%$ |
| **TapNet** (Ours, Conv4) | $\mathbf{50.68 \pm 0.11\%}$ | $\mathbf{69.00 \pm 0.09\%}$ |
| **Relation Networks** | $50.44 \pm 0.82\%$ | $65.32 \pm 0.70\%$ |
| **Transductive Propagation Nets** | $\mathbf{55.51 \pm 0.86\%}$ | $\mathbf{69.86 \pm 0.65\%}$ |
| **SNAIL** | $55.71 \pm 0.99\%$ | $68.88 \pm 0.92\%$ |
| **adaResNet** | $56.88 \pm 0.62\%$ | $71.94 \pm 0.57\%$ |
| **TapNet** (Ours, ResNet-12-small) | $\mathbf{59.47 \pm 0.12\%}$ | $\mathbf{72.79 \pm 0.10\%}$ |
| **TADAM-$\alpha$** | $56.8 \pm 0.3\%$ | $75.7 \pm 0.2\%$ |
| **TADAM-TC** | $58.5 \pm 0.3\%$ | $\mathbf{76.7 \pm 0.3\%}$ |
| **TapNet** (Ours, ResNet-12) | $\mathbf{61.65 \pm 0.15\%}$ | $76.36 \pm 0.10\%$ |

*Table 4.* Few-shot classification accuracies for 5-way *tiered*ImageNet

| Methods | 5-way *tiered*ImageNet | |
|---|---|---|
| | 1-shot | 5-shot |
| **MAML** | $51.67 \pm 1.81\%$ | $70.30 \pm 1.75\%$ |
| **Prototypical Nets** | $53.31 \pm 0.89\%$ | $72.69 \pm 0.74\%$ |
| **Relation Nets** | $54.48 \pm 0.93\%$ | $71.31 \pm 0.78\%$ |
| **Transductive Propagation Nets** | $\mathbf{59.91 \pm 0.94\%}$ | $73.30 \pm 0.75\%$ |
| **TapNet** (Ours, Conv4) | $57.11 \pm 0.12\%$ | $\mathbf{73.66 \pm 0.09\%}$ |

second option is a smaller version of ResNet-12, like the one used in SNAIL. With this version of ResNet-12, each residual block is constructed with three $3 \times 3$ convolutional layers. Also, a $1 \times 1$ convolutional layer is used for residual connection for each block. This network consists of four residual blocks with 64, 96, 128 and 256 respective channels. We denote this network "ResNet-12-small" in comparison to the larger version of ResNet-12 in the main paper. The Adam optimizer is also used for optimizing for both Conv4 and ResNet-12-small.

We focus on 5-way miniImageNet classification and the measured accuracy results are presented in Table 3. The few-shot learners are shown in four groups, depending on the required base network size. We notice significant performance differences in general as the model size/complexity changes. The first group uses the Conv4 network. Among the methods here, our TapNet achieves best 1-shot and 5-shot accuracies. The methods in the second group, Transductive Propagation Networks (TPN) and Relation Networks, are also based on the Conv4 embedder, but require additional networks for certain purposes. TPN utilizes an additional convolutional block in the graph construction network, and Relation Networks use additional convolutional blocks in

its relation module. Both these methods require significant extra learning efforts, compared to the learners relying mostly on the Conv4 base embedder. TPN achieves higher 1-shot and 5 shot accuracies than the methods in the first group. The next group of methods uses ResNet-12-small. SNAIL and adaResNet are compared with TapNet. TapNet again achieves the best 1-shot and 5-shot performance in this group. The methods in the last group utilize ResNet-12, which is the largest network among the feature extractors considered in this work. For 1-shot results, our TapNet once again provides the best accuracy. For 5-shot, TapNet's result is comparable to that of the best method, TADAM-TC, in the sense that the confidence intervals overlap. In summary, given the same base network, TapNet consistently gives either the best accuracy or one comparable to the best in 5-way *mini*ImageNet classification among the well-known methods.

In Table 4, we display additional results on 5-way *tiered*ImageNet classification with Conv4 embedding. In the *tiered*ImageNet experiment, we had a small modification to the embedding network in TapNet. We added a $2 \times 2$ average pooling layer on top of the Conv4 network. Also, for 1-shot *tiered*ImageNet classification we found that it

Table 5. Number of parameters required for the learners

| Method | Feature extractor | Additional Conv layer | Additional learnable parts |
|---|---|---|---|
| **Matching Networks** | 112k | - | - |
| **Prototypical Networks** | 112k | - | - |
| **MAML** | 112k | - | 12k (FC layer) |
| **TapNet** (Conv4) | 112k | - | 46k (Parameters of $\Phi$) |
| **Relation Nets** | 112k | 111k | 4k (FC layer) |
| **Transductive Prop. Nets** | 112k | 37k (Graph Construction) | - |
| **SNAIL** (ResNet-12-small) | 2.2M | 1.3M + $\alpha$ (Attention + TC) | - |
| **adaResNet** | 2.2M | 0.3M | - |
| **TapNet** (ResNet-12-small) | 2.2M | - | 5k (Parameters of $\Phi$) |
| **TADAM**-$\alpha$ | 9.4M | - | - |
| **TADAM**-TC | 9.4M | - | 1.2M (FC layer) |
| **TapNet** (ResNet-12) | 9.4M | - | 10k (Parameters of $\Phi$) |

was beneficial to use the higher-shot training strategy; we adopted 4-shot meta-training for 1-shot classification. As a result, TapNet achieves the best accuracy for 5-shot classification, and the second best accuracy for 1-shot among the methods using the same Conv4 embedding network.

## 4. Number of Network Parameters

It would be useful to understand the required complexity levels of the learning methods compared in this work. We in particular look at the number of network parameters used in each method. We focus on the number of parameters for the convolutional layer and other important learnable parts which are directly related to the learning efforts of the network. In particular, the other learnable parts include the fully connected (FC) layers in some cases and the stand-alone linear weights for the class reference vectors in Tap-Nets. The Conv4 network requires 112,320 parameters in total. The residual networks rely on considerably larger numbers of learnable parameters. ResNet-12-small runs on 2.2 million parameters, while ResNet-12 requires 9.4 million parameters approximately. Table 5 includes the number of parameters necessary for implementing any additional convolutional layers or learnable parts for each method. In the case of SNAIL, we marked the number of additional parameters simply as $\alpha$, since the numbers of parameters used for attention blocks and temporal convolution blocks there are hard to estimate due to the lack of available detail descriptions. We note that for the same number of parameters, the convolutional layer, which utilizes a sliding window to repeat many multiply/add operations, requires substantially higher computational complexity than the other types of learnable parts considered in Table 5.



\end{document}